\documentclass[10pt,twocolumn,letterpaper]{article}

\usepackage{iccv}
\usepackage{times}
\usepackage{epsfig}
\usepackage{graphicx}
\usepackage{amsmath}
\usepackage{amssymb}

\usepackage{booktabs}
\usepackage{makecell}
\usepackage{newfloat}
\usepackage{caption}
\usepackage{multirow}
\usepackage{tablefootnote}
\usepackage{comment}
\usepackage{algorithm}
\usepackage{algpseudocode}
\usepackage[accsupp]{axessibility}  % Improves PDF readability for those with disabilities.
\usepackage[breaklinks=true,bookmarks=false]{hyperref}

\iccvfinalcopy % *** Uncomment this line for the final submission

\def\iccvPaperID{2745} % *** Enter the ICCV Paper ID here

\ificcvfinal\fi

\begin{document}

%%%%%%%%% TITLE
\title{SCANet: Scene Complexity Aware Network for Weakly-Supervised Video Moment Retrieval}

\author{Sunjae Yoon \hspace{0.5cm} Gwanhyeong Koo \hspace{0.5cm} Dahyun Kim \hspace{0.5cm} Chang D. Yoo\\
Korea Advanced Institute of Science and Technology (KAIST)\\
{\tt\small \{sunjae.yoon,kookie,dahyun.kim,cd\_yoo\}@kaist.ac.kr}
}
% For a paper whose authors are all at the same institution,
% omit the following lines up until the closing ``}''.
% Additional authors and addresses can be added with ``\and'',
% just like the second author.
% To save space, use either the email address or home page, not both
%\and
%Second Author\\
%Institution2\\
%First line of institution2 address\\
%{\tt\small secondauthor@i2.org}
%}

\maketitle
% Remove page # from the first page of camera-ready.
\ificcvfinal\thispagestyle{empty}\fi

%Video moment retrieval aims to localize moments in video corresponding to a given language query. To avoid the expensive cost of annotating the temporal movements for performing this task, weakly-supervised video moment retrieval (wsVMR) systems have been studied. For such systems, generating a number of proposals as moment candidates and then selecting the most appropriate candidate has been a popular approach. These proposals that are assumed to contain distinguishable scenes in a video as candidates are heuristically determined irrespective of the video.  To improve performance, the proposed retrieval system referred to as Scene Complexity Aware Network (SCANet) evaluates the scene complexity of the video in generating the proposals. Experimental results on three retrieval benchmarks (i.e., Charades-STA, ActivityNet, TVR) achieve state-of-the-art performances and demonstrate the effectiveness of incorporating the scene complexity.
%
%Although these proposals are assumed to contain distinguishable scenes in a video as candidates, existing proposal generation methods are heuristically determined irrespective of the number of scenes in a video.
%
%%%%%%%%% ABSTRACT
\begin{abstract}
   Video moment retrieval aims to localize moments in video corresponding to a given language query. To avoid the expensive cost of annotating the temporal moments, weakly-supervised VMR (wsVMR) systems have been studied. For such systems, generating a number of proposals as moment candidates and then selecting the most appropriate proposal has been a popular approach. These proposals are assumed to contain many distinguishable scenes in a video as candidates. However, existing proposals of wsVMR systems do not respect the varying numbers of scenes in each video, where the proposals are heuristically determined irrespective of the video. We argue that the retrieval system should be able to counter the complexities caused by varying numbers of scenes in each video. To this end, we present a novel concept of a retrieval system referred to as Scene Complexity Aware Network (SCANet), which measures the `scene complexity' of multiple scenes in each video and generates adaptive proposals responding to variable complexities of scenes in each video. Experimental results on three retrieval benchmarks (\ie Charades-STA, ActivityNet, TVR) achieve state-of-the-art performances and demonstrate the effectiveness of incorporating the scene complexity. The project is available at: \href{https://github.com/dbstjswo505/SCANet}{\texttt{github.com/dbstjswo505/SCANet}}
\end{abstract}

%%%%%%%%% BODY TEXT
\section{Introduction}
\label{sec:intro}
% VMR의 필요성, 문제점
Video search has the core building block of recently growing video streaming services (\eg YouTube, Netflix).
%
%Video searching technologies have been the essential building blocks of recently growing video streaming services (\eg YouTube, Netflix). 
%
To enhance the capability of video search, video moment retrieval (VMR) aims to localize the start and end time of the moment pertinent to a given language query in an untrimmed video.
%
%To enhance the capability of video search, video moment retrieval (VMR) aims to localize the start and end time of the moment pertinent to a given language query in an untrimmed video.
%
%
The success of the VMR provides us with accurate video contextual information in less time and effort. 
Until recently, these remarkable search performances have been dependent on the size and quality of labeled training datasets.
%
% has been dependent on the size and quality of  labelled training dataset.
%
However, these datasets cost a labor-intensive annotating process (\ie Annotators should find the start-end time of moments corresponding to query descriptions), and sometimes the annotated moments are ambiguous.
% wsVMR 등장 및 wsVMR의 문제점
To cope with this problem, many weakly-supervised VMR (wsVMR) methods \cite{mithun2019weakly,duan2018weakly,zhang2020counterfactual,zheng2022weakly2} have been proposed by only utilizing the video-query pairs, which are less laborious to annotate.
%
\begin{comment}
\begin{figure}[t]
  \centering
  \includegraphics[width=\linewidth]{./figs/intro.pdf}
   \caption{Scene-proposal mismatch problem of current systems resulted from spurious correlations between video length and the number of proposals: (a) shows an unnecessary bunch of proposals on a long-length video containing single scene and (b) shows scarce proposals on a short-length video containing many scenes.}
   \label{fig:intro1}
\end{figure}
\end{comment}

To perform the weak supervision using video-query pairs, if one query is paired (\ie annotated) with multiple videos, we can identify the common scene among these videos and determine the alignment between the query and the scene.
To implement this, all videos are divided into multiple segments, and the retrieval system maximizes the similarity scores between each query and paired segments while suppressing the scores between the query and unpaired segments in other videos.
During the inference, the system selects a segment with the highest score as a moment prediction for a given query.
For the wsVMR systems to accurately classify the best segment in a video, numerous video-language joint representation learning methods \cite{lin2020weakly,song2020weakly,zheng2022weakly2,wang2021visual} have been proposed.
\begin{figure*}[t]
  \centering
  \includegraphics[width=\linewidth]{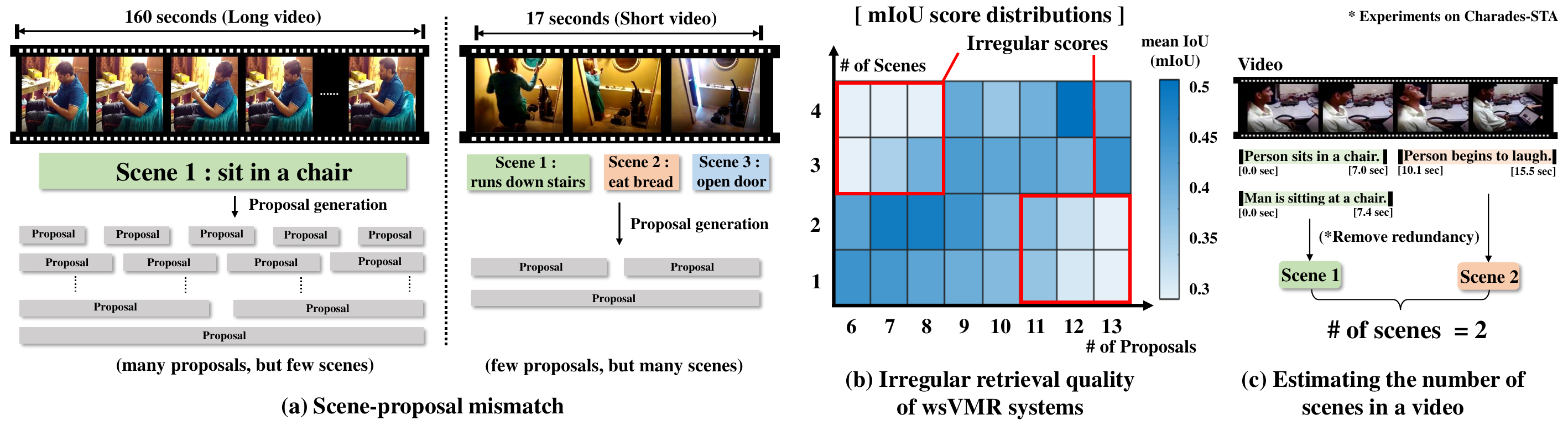}
   \caption{Scene-proposal mismatch in current wsVMR systems: (a) shows an unnecessary many proposals on a video containing few scenes and few proposals on a video containing many scenes, (b) shows mIoU scores of the current model's predictions according to the number of scenes and the number of generated proposals and (c) shows a method for estimating the number of scenes, where redundant scenes are removed from the counts.}
   \label{fig:intro1}
   \vskip -0.1in
\end{figure*}

Recently, researchers also have another focus on a study of how to generate video segments to capture many scenes in a video \cite{ma2020vlanet,zheng2022weakly2}.
%Recently, researchers also have another focus on a study of how to generate these video segments of minimum to capture all scenes in a video \cite{ma2020vlanet,zheng2022weakly2}.
%
These segments are referred to as `candidate moment proposals', which is crucial, as they directly affect the retrieval performances by regulating the proposal quantities.
Unfortunately, as supervision is not available in generating proposals, wsVMR systems \cite{zheng2022weakly1,zheng2022weakly2} use a fixed number of proposals for all input videos 
under heuristic optimization of a specific dataset, which is not reasonable to deal with varying numbers of scenes in a video.
%
%Generating
%[chk]
While some methods \cite{ma2020vlanet,huang2021cross} consider varying numbers of proposals for each video, they still rely on spurious correlations, such as generating proposals proportionally to the video length or using sliding window.
Therefore, the current proposal generation method could not accurately respond to the diverse number of scenes in each video. 
We refer to this situation as a `scene-proposal mismatch'.
%따라서 현재의 proposal 생성 방식은 비디오 마다 다른 장면들의 양에 적적히 대응할 수가 없었다.
%비디오에 실제로 얼마나 많은 후보 장면들이 있는지 그리고 생성된 proposal양의 그것에 잘 대응할 수 있는지에 대한 고민이 없었다.
%
%Therefore current proposal generations do not how many scenes are in the video.
%properly capture the scenes in a video.
%
%Irregular retrieval quality of current wsVMR systems, which results in `scene-proposal mismatch' that does not properly capture the scenes in a video.
%
%For an instance, Figure \ref{fig:intro1} presents this scene-proposal mismatch of current systems \cite{ma2020vlanet,huang2021cross} that spuriously correlated with video lengths. 
%
%
For instance, in Figure \ref{fig:intro1}(a), the systems produce an unnecessarily large number of proposals by referring to the long length of the video, but the video only contains a single scene (\ie scene of sitting still in a chair throughout the video), which should be handled by small amounts of proposals. 
They also show scene-proposal mismatch by producing a small number of proposals for the video containing many scenes, such that those scarce proposals seem not to work correctly.
%few는 그림에 있는거라서 되도록이면 사용하고자함.
%그림을 바꿔야겠음

%
Our experimental evidence in Figure \ref{fig:intro1}(b) validates the current wsVMR systems' incorrectness due to the scene-proposal mismatch.
We plot performances (\ie mean Intersection over Union (mIoU) scores) over predictions along the number of scenes in videos and the number of proposals generated, which shows irregularities in the scores. 
The scores are low for videos with many scenes but few proposals and also low for videos with few scenes but many proposals.
To estimate the number of scenes in a video, as shown in Figure \ref{fig:intro1}(c), we counted the number of paired queries for each video as a discrete approximation of the scene. 
Here, we found that some queries describing the same scene led to redundancy in the counting.
Thus, we remove the redundancy of those queries via calculating their IoUs\footnote{We remove redundancy by scenes with IoU $>$ 0.5.} between temporal boundary annotations\footnote{Temporal annotations are used only for identifying the proposal-scene mismatch problem and they are not involved in the wsVMR task}.
Our study further showed that the scene-proposal mismatch affects about 41\% of videos in VMR benchmarks (\ie Charades-STA \cite{gao2017tall}, ActivityNet-Caption \cite{krishna2017dense}).
%
%One straightforward solution is to perform frame-level moment prediction \cite{zhang2020learning,wang2021structured} without using candidate proposals.
%
%However, as the frame-level supervisions (\ie temporal annotation) are not provided in weakly-supervised settings, training is not available, or it requires pseudo-label, which can be ad-hoc and minor solutions for specific data.
%

Intrigued by the scene-proposal mismatch, this paper proposes a wsVMR system referred to as Scene Complexity Aware Network (SCANet), which allows the system to mitigate the scene-proposal mismatch problem and generate proposals adaptive to the complexity of the scenes contained in the video.
%the variance of scene quantity for each video.
%
For a given input video, SCANet first defines the scene complexity with a scalar, meaning how difficult for the system to find (\ie retrieve) a specific scene among multiple distinguishable scenes in the video, which can be effective prior knowledge of video by complementing weak supervision of VMR.
On top of the scene complexity, SCANet adaptively generates proposals and enhances their representations.
Therefore, SCANet incorporates (1) Complexity-Adaptive Proposal Generation (CPG) that generates adaptive proposals by leveraging the quantities of proposals under consideration of the complexity and (2) Complexity-Adaptive Proposal Enhancement (CPE) that enhances the proposals' representations corresponding to the scene complexity.
%%
%Here, some redundant queries annotated to the same scene are removed by inspecting word-level overlaps among the queries.
%
Furthermore, motivated by recent success \cite{zhang2020counterfactual,zheng2022weakly2} of contrastive learning for wsVMR system, we introduce technical contributions to mine hard negatives in the input video and further video corpus together under our designed framework.
%The SCANet incorporate the scene complexity, where CPG generates candidate proposals by deciding their amounts and lengths, and CPE performs reconstruction and contrastive learning to make alignment between the proposals and query, where the training losses are calibrated by scene complexity.
%
Our extensive experiments show the effectiveness of the proposed SCANet, and qualitative results validate enhanced interpretability.
%시스템의 비디오 내 장면들에 대한 학습 난이도를 추정하는것을 목표호 한다.
%
%SCANet incorporates (1) Query-guided Scene Complexity Estimator (QCE), which reasons the number of scenes as the scene complexity for a given video via paired queries' priors, and (2) Complexity-Adaptive Proposal Generation (CPG), which regulates proper quantities of candidate proposals via considering the estimated scene complexity, and (3) Complexity-Adaptive Representation Learning (CPE), which enhance the joint representation between query and proposals via incorporating multiple representation learning frameworks and selectively weighting them corresponding to the scene complexity.
%
%In the overall pipeline, for given a query and video, QCE estimates the scene complexity of the video by referring to the number of annotated queries.
%
%Here, some redundant queries annotated to the same scene are removed by inspecting word-level overlaps among the queries.
%
%Founded on the scene complexity, CPG generates candidate proposals by deciding their amounts and lengths, and CPE performs reconstruction and contrastive learning to make alignment between the proposals and query, where the training losses are calibrated by scene complexity.
%
%-------------------------------------------------------------------------
\begin{figure*}[t]
  \centering
  \includegraphics[width=\linewidth]{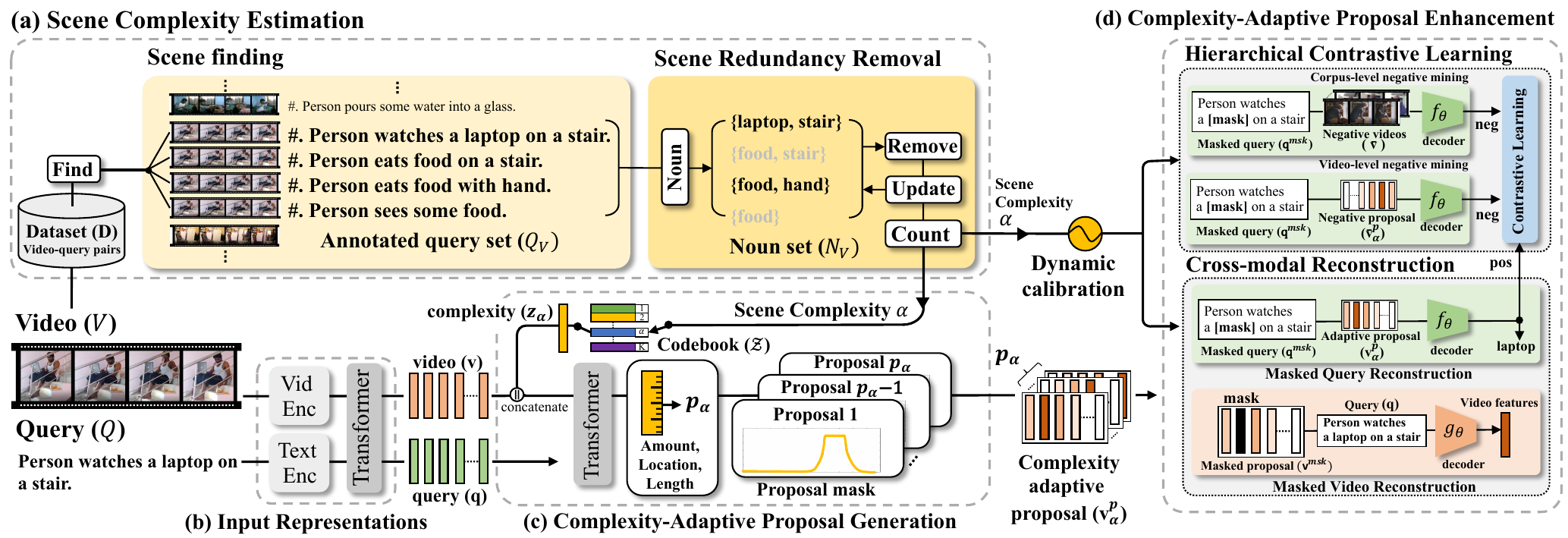}
   \caption{Illustration of proposed SCANet. (a) shows a scene complexity estimation which takes an input video and estimates a scene complexity using video-query pairs, (b) shows input representations, (c) shows a complexity-adaptive proposal generation which generates adaptive proposals according to the complexity, and (d) shows a complexity-adaptive proposal enhancement, which introduces multiple representation enhancements and calibrates them corresponding to the complexity.}
   \label{fig:model}
\end{figure*}
\section{Related Works}
\subsection{Advancements in Video Moment Retrieval}
%비디오 검색기술은 사용자의 자유로운 
%
Video Moment Retrieval (VMR) \cite{gao2017tall}, as one of the high-level vision-language tasks, aims to localize video segments corresponding to scene descriptions automatically.
Previous successes of multi-modal interaction \cite{vaswani2017attention,lu2019vilbert} have contributed to many respectful works \cite{yuan2019semantic,zeng2020dense,wang2020dual,zhang2020learning,liu2022umt} to boost retrieval performances by improving the joint representation of video and language to understand their semantic similarities.
%
%
%TALL \cite{gao2017tall} has been the first attempt to perform VMR, where it integrates video and language query into joint spaces and predicts the start-end time of the video that describes the query.
%
SMIN \cite{wang2021structured} and MPGP \cite{sun2022you} show the recent state-of-the-art performances of video searching technologies.
Researchers are also challenged to make these VMR systems to be more generalized and practical.
For general usage of VMR, corpus-level retrieval systems \cite{lei2020tvr,li2020hero,yoon2022selective,yoon2023counterfactual} have been proposed, and for practical usage, fast retrieval system \cite{gao2021fast} has been proposed.
%VMR 성장 3~4줄
Moreover, the laborious annotating problem has been another historical issue of VMR systems.
Annotating video segments corresponding to given scene descriptions is quite difficult and sometimes inaccurate due to temporal ambiguity.
To overcome this, weakly-supervised learning methods have been considered, where it is assumed that systems are not given temporal annotations (start-end time).
There has been much literature on weakly-supervised methods. 
We elaborate on this below in another section with detailed explanations.
\subsection{Weakly-supervised Video Moment Retrieval}
Weakly-supervised Video Moment Retrieval (wsVMR) shares the same goal as VMR and also aims to reduce the cost of annotation.
Therefore, researchers have made an effort to train video-language alignment without temporal boundary annotations, where they considered introducing more affordable supervision.
%
%%TGA: First video-sentence aligning
%WS-DEC: First
TGA \cite{mithun2019weakly} and WS-DEC \cite{duan2018weakly} were the first weakly-supervised VMR systems that utilized the pairing information in video-language pairs\footnote{Fully-supervised setting uses moment-language pairs for training.} as a weak-supervision for the alignment.
Annotating video-language pairs is less laborious than moment-language pairs for fully-supervised learning. Thus many works have been performed in this weakly-supervised setting.
%
%% 1 문장
%[d. multi modal attention]
%[d. multi modal attention] VLANet: Vide-Language aligment Cascaded Attention
%[d. multi modal attention] LoGAN: Graph Attention
%%[a. moment prediction refining]
%[a. moment prediction refining] WSTAN: weakly-supervised 2D score map
%[a. moment prediction refining] CTF: Coarse-To-Fine
%[a. moment prediction refining]: LCNet fine-grained correspondences between video and text for temporal sentence grounding by using self-supervised learning.
%[a. moment prediction refining] CRM: Cross-Sentence Relation
%[a. moment prediction refining]: RTBPN
To improve multi-modal interactions, attention-based models \cite{ma2020vlanet,tan2021logan} have been proposed. In addition, to achieve fine-grained retrieval, methods for refining predictions \cite{wang2021weakly,yoon2021weakly,chen2020look,yang2021local,huang2021cross,zhang2020regularized,wu2020reinforcement} have also been developed.
These methods have made significant contributions to generating candidate moment proposals to predict.
%To enhance the multi-modal interactions, attention-based models \cite{ma2020vlanet,tan2021logan} were proposed, and to perform fine-grained retrieval, prediction refining methods \cite{wang2021weakly,chen2020look,yang2021local,huang2021cross,zhang2020regularized,wu2020reinforcement} have also been proposed, where they further made significant contributions in candidate moment proposal generations for the moment prediction.
%To improve multi-modal interactions, attention-based models \cite{ma2020vlanet,tan2021logan} have been proposed. In addition, to achieve fine-grained retrieval, methods for refining predictions \cite{wang2021weakly,chen2020look,yang2021local,huang2021cross,zhang2020regularized,wu2020reinforcement} have also been developed
%%[c. reinforcement learning]
%[c. reinforcement learning] BAR: Boundary Adaptive Refinement (BAR) framework that resorts to reinforcement learning (RL) to guide the process of progressively refining the temporal boundary.
%Reinforcement learning \cite{wu2020reinforcement} is also available to perform progressive refinement in moment predictions.
%
%%[b. reconstruction]
%[b. reconstruction] SCN: Semantic Complemtion
%[b. reconstruction] MARN: Multi-level Attentional Reconstruction
%[b. reconstruction] EC-SL
With the success of self-supervised learning, recent wsVMR systems \cite{chen2021towards,lin2020weakly,song2020weakly} introduce the word reconstruction framework from the masked word in the query sentence. 
%%[e. hard negative mining for contrastive leaning]
%[e. hard negative mining] WSRA:
%[e. hard negative mining] CCL
%[e. hard negative mining] CNM
%[e. hard negative mining] CPL
%[h. positive sample mining]
%[h. positive sample mining] VCA
%[h. positive sample mining] WSLLN
Henceforth, contrastive learning achieves large performance gains via mining hard negative retrievals \cite{fang2020weak,zhang2020counterfactual,zheng2022weakly1,zheng2022weakly2} and positive retrievals \cite{wang2021visual,gao2020weakly}.
However, current systems have never considered the scene-proposal mismatch problem and still suffer from this. Thus we first propose a method to mitigate the mismatch via scene complexity measurements.
%

%The scene complexity determines the difficulty of selecting (or retrieving) a specific scene among multiple scenes in a given video. It is an effective prior knowledge that can be incorporated into weak supervision.
\section{Method}
Figure \ref{fig:model} presents an overall pipeline of the proposed Scene Complexity Aware Network (SCANet) for retrieval systems.
SCANet first takes a video and measures scene complexity by estimating how many different scenes are in the video.
The scene complexity determines the difficulty of selecting (\ie retrieving) a specific scene among multiple scenes in a given video, which is effective prior knowledge that can be incorporated into weak supervision.
Founded on the scene complexity, SCANet builds (1) Complexity-Adaptive Proposal Generation (CPG) and (2) Complexity-Adaptive Proposal Enhancement (CPE).
The CPG adaptively leverages proposal generation, which mitigates the scene-proposal mismatch, and the CPE enhances the proposals' representations and dynamically calibrates enhancements according to the scene complexity.
%
%
%
%In inference, wsVMR system under ISM framework selects the most appropriate proposal among the generated proposal candidates.
%
\subsection{Scene Complexity}
Videos contain a varying number of scenes, and if we can know about the quantities of scenes existing in each video, it should be an effective prior knowledge by giving specified search space to perform retrieval in the video (\ie especially effective in the method of generating retrieval candidates like moment proposals).
In that sense, our proposed scene complexity aims to make the retrieval system identify how many scenes exist in the search space of a given video.
To this end, we propose a scene complexity estimation algorithm, which takes inputs from a video $V$ and the dataset $D$ composed of video-query pairs and produces the number of scenes contained in the video as given below:
\begin{eqnarray}
\alpha = f_{sc}(V,D) \in \mathbb{R}^{1},
\end{eqnarray}
where $\alpha$ is the number of different scenes in the video and we define it as the scene complexity. 
Following, we present a detailed process of the $f_{sc}$, which includes two procedures: (1) scene finding and (2) scene redundancy removal. 
\paragraph{Scene finding.} Scene finding is to specify all the candidate scenes in a given video. 
To this, we utilized the annotated queries sharing the same video as a discrete approximation of the scenes.
For obtaining the annotated queries, we investigate video ID\footnote{Previously, video ID (\eg `ID: 0BH84') is just used for accessing the feature data, but we further utilize the ID to build $Q_{V}$.} and collect the queries that share the same video ID among the video-query pairs dataset $D$:
\begin{eqnarray}
Q_{V} = \textrm{Find}(V_{id},D),
\end{eqnarray}
where $V_{id}$ is video ID, $Q_{V}$ is annotated query set and $\textrm{Find}(\cdot,\cdot)$ is a function to collect queries sharing the same video ID.
Figure \ref{fig:model}(a) also gives examples of annotated queries, but we also find a semantic redundancy\footnote{video-query pair datasets usually contain many redundant queries} among the queries (\eg queries meaning ``eating food'').
It is required to remove the redundancy for the accurate counting of the number of different scenes.
Therefore we devise a redundancy removal for our scene complexity estimation.

\paragraph{Scene Redundancy Removal.} We utilize the word overlaps\footnote{Unlike Figure 1(c), labels (start-end times) are unavailable in wsVMR.} to sense the semantic redundancy among the queries in the annotated query set $Q_{V}$.
In detail, we identify a part of speech of all words in the queries and sample nouns using the natural language toolkit \cite{loper2002nltk} and then remove the redundancy from the queries that have overlaps by a noun.
An example of this process is shown in Figure \ref{fig:model}(a), where the annotated query set is provided as $Q_{V} =$\{`Person watches a laptop on a stair', `Person eats food on a stair', `Person eats food with hand', `Person sees some food'\}, we should exclude the redundant queries that share similar meaning.
To estimate which queries are semantically redundant, we identify nouns in each query as $N_{V}=$\{\{laptop, stair\}, \{food, stair\}, \{food, hand\}, \{food\}\}, where $N_{V}$ is defined as annotated noun set. 
Here, nouns meaning human (\eg person) are excluded from the findings.
If there is word overlap between elements of $N_{V}$, the element with the most overlapping is removed first.
This process is continued until there is no overlap.
Thus, the redundancy removal process prunes out the elements in $N_{V}$ and updates upto $N_{V}=$\{\{laptop, stair\}, \{food, hand\}\} or \{\{laptop, stair\}, \{food\}\} after that, we count the number of elements in $N_{V}$ as the final number of scenes in the given video $V$.
To summarize the scene complexity estimation process, we formally define an algorithm about $f_{sc}$ below:
\begin{algorithm}[ht]
    \begin{algorithmic}[1]
	\State \textbf{Input}: Video $V$, video-query pairs dataset $D$
	\State \textbf{Output}: Scene complexity $\alpha$
	\State Find annotated query set: $Q_{V} = \text{Find}(V_{id},D)$ 
	\State Find nouns: $N_{V} = \text{Noun}(Q_{V})$
	\State \textbf{while} $\exists \text{word overlap among elements in $N_{V}$}$  \textbf{do}
	%\State \textbf{for} $i \leftarrow 1 $ \textbf{to} $T$ \textbf{do}
	\State \quad Remove the most overlapping:  $N_{V} \leftarrow \text{Remove}(N_{V})$
	\State \textbf{end}
	\State \textbf{return} The number of elements of $N_{V}$
	\end{algorithmic}
	\caption{Scene complexity estimation algorithm $f_{sc}$}
	\label{alg:Cascaded_mpg}
\end{algorithm}

\noindent Noun($\cdot$) is a function to filter out noun\footnote{Table 4 gives ablation studies (\eg verb) to identify the redundancy.} and Remove($\cdot$) is a function to find the element with the most overlap and remove it.
The final number of elements in $N_{V}$ is defined as scene complexity $\alpha$ of an integer scalar.
Following, our proposed SCANet utilizes $\alpha$ to adapt the retrieval in terms of proposal generation and proposal enhancement.

\subsection{Scene Complexity Aware Network}
\paragraph{Input Representations.}
We first give formal definitions of the input video $V$ and query $Q$ used in SCANet.
%
%SCANet takes a video and query and embeds them into $d$-dimensional space.
%
Frame-level video features are obtained from a pre-trained video encoder \cite{carreira2017quo,tran2015learning} and word-level query features are obtained from text encoder \cite{pennington2014glove}.
Both features are embedded into $d$-dimensional joint space. 
After adding positional encoding \cite{vaswani2017attention} and applying layer normalization \cite{ba2016layer}, we get the final video features $\mathbf{v} \in \mathbb{R}^{N_{v} \times d}$ and the query features $\mathbf{q} \in \mathbb{R}^{N_{q} \times d}$, where $N_{v}$ is the number of video frames and $N_{q}$ is the number of words in the query.
\paragraph{Multi-Modal Interaction.}
To give multi-modal interactions between the query and video, we use Transformer Attention \cite{vaswani2017attention}. 
The video features $\mathbf{v}$ and query features $\mathbf{q}$ are concatenated and prepared for the Attention inputs below:
\begin{eqnarray}
 [\mathbf{v}||\mathbf{q}] = \text{Attention}([\mathbf{v}||\mathbf{q}]) \in \mathbb{R}^{(N_{v}+N_{q}) \times d},
 \label{eq:multimodal}
\end{eqnarray}
where [$\cdot||\cdot$] denotes the concatenation and we get attended video features $\mathbf{v} \in \mathbb{R}^{N_{v} \times d}$ and query features $\mathbf{q} \in \mathbb{R}^{N_{q} \times d}$.
\subsection{Complexity-Adaptive Proposal Generation}
Complexity-Adaptive Proposal Generation (CPG) in Figure \ref{fig:model}(c) is designed to generate candidate moment proposals adapting the scene complexity $\alpha$  of a given video, where $\alpha$ accounts for three aspects of proposals: amount, location, and length.
To implement this, we devise a `complexity vector' corresponding to the complexity level, such that we build a codebook $\mathcal{Z}=\{\mathbf{z}_{k}\}_{k=1}^{K} \in \mathbb{R}^{K \times d}$ composed of $d$-dimensional learnable $K$ vectors and select a single vector by indexing $\alpha$ as $\mathbf{z}_{\alpha} \in \mathbb{R}^{d}$, where the $K$ (\eg $K=8$) is the maximum number of $\alpha$.
Thus the $\mathbf{z}_{\alpha}$ is our defined complexity vector that has sensibility according to the $\alpha$. 
After providing semantics of input modalities as $[\mathbf{z_{\alpha}}||\mathbf{v}||\mathbf{q}] = \textrm{Attention}([\mathbf{z_{\alpha}}||\mathbf{v}||\mathbf{q}])$, in the following, the $\mathbf{z}_{\alpha}$ generates adaptive proposals deciding proposal properties in terms of amount, location and length.
\begin{comment}
\begin{eqnarray}
 [\mathbf{z_{\alpha}}||\mathbf{v}||\mathbf{q}] = \text{Attention}([\mathbf{z_{\alpha}}||\mathbf{v}||\mathbf{q}],[\mathbf{z_{\alpha}}||\mathbf{v}||\mathbf{q}],[\mathbf{z_{\alpha}}||\mathbf{v}||\mathbf{q}]).
 \label{eq:multimodal2}
\end{eqnarray}
\end{comment}
%

To decide the amount of proposals, we first build an integer set $I=\{p_{\textrm{min}},p_{\textrm{min}}+1\cdots,p_{\textrm{max}}\} \in \mathbb{R}^{n}$ regarding the number of proposals, where $p_{\textrm{min}}$ and $p_{\textrm{max}}$ are the minimum (\eg 5) and the maximum number (\eg 10) of proposals, and the $n$ = $p_{\textrm{max}} - p_{\textrm{min}} + 1$ is the number of elements in the $I$.
The complexity vector $\mathbf{z}_{\alpha}$ decides a single number $p_{\alpha} \in I$ from the integer set $I$ and generates $p_{\alpha}$ proposals.
For the detailed implementation of this, we utilize a Multi-Layer Perceptron (MLP) that takes the input of $\mathbf{z}_{\alpha}$ and produces an $n$-dimensional selection vector as $\mathbf{a} = \text{MLP}(\mathbf{\mathbf{z}_{\alpha}}) \in \mathbb{R}^{n}$, where the $n$ denotes the same dimension of the integer set $I$.
The selection is performed by using an output of argmax($\mathbf{a}$) as index to select a single integer in $I$ and the selected integer defines the number of proposals $p_{\alpha}$.
To make selection trainable, we introduce a Gumbel-Softmax \cite{jang2016categorical}, which provides functional $n$-dimensional one-hot vector $\mathbf{g}$ for the deciding the number of proposals over integer set $I$ below:
\begin{equation}
\begin{aligned}
 \mathbf{g} &= \text{Gumbel\_Softmax}(\mathbf{a}) \in \mathbb{R}^{n},\\
 p_{\alpha} &= \sum_{i=1}^{n}\mathbf{g}_{i} \cdot I_{i} \in \mathbb{R}^{1},
\end{aligned}
\end{equation}
where the number of proposals (\ie amount of proposals) is finally determined by the number $p_{\textrm{min}} \leq p_{\alpha} \leq p_{\textrm{max}}$.

To decide the locations and lengths of proposals, we first build a proposal mask, which remains the video features in the region of the mask as the proposal features.
A center point and width of the operating region of the mask correspond to the location and length of the proposal, thus complexity features $\mathbf{z}_{\alpha} \in \mathbb{R}^{1 \times d}$ regresse the center and width as $[\mathbf{c}_{\alpha},\mathbf{w}_{\alpha}] = \sigma(\mathbf{z}_{\alpha}W_{p}) \in \mathbb{R}^{2}$, where $W_{p} \in \mathbb{R}^{d \times 2}$ is learnable weights for regression $\sigma(\cdot)$ is the sigmoid function, and $\mathbf{c}_{\alpha} \in \mathbb{R}^{1}$, $\mathbf{w}_{\alpha} \in \mathbb{R}^{1}$ are the center and width of the proposal.

Founded on the $\mathbf{c}_{\alpha}$ and $\mathbf{w}_{\alpha}$, we design the proposal mask, and here, Figure \ref{fig:mask}(a) shows a popular example of the proposal mask using Gaussian curve \cite{zheng2022weakly1,zheng2022weakly2}.
We consider that the Gaussian curve may not be reasonable because video features are unevenly attended inside the proposal. 
Therefore, as shown in Figure \ref{fig:mask}(b), we design Flatten Gaussian mask, which is simple, yet more reasonable by evenly remaining features inside the proposal\footnote{Table 5 validate also the effectiveness of Flatten Gaussian curve.}.
To give a formal definition of our mask, we first construct the base mask using the Gaussian curve using the $\mathbf{c}_{\alpha}$ and $\mathbf{w}_{\alpha}$ as given below:
%
%Together with the number of proposals $P$, the final mask $\mathbf{m}_{p} \in \mathbb{R}^{N_{v}}$ for Complexity-Adaptive Proposal Generation is defined:
%
\begin{figure}[t]
  \centering
  \includegraphics[width=\linewidth]{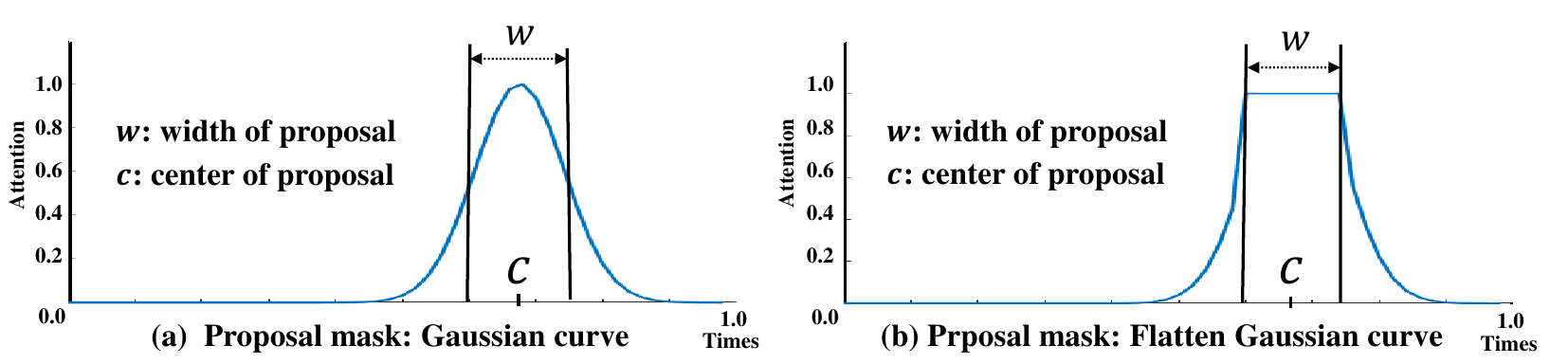}
   \caption{Illustration of (a) gaussian mask \cite{zheng2022weakly1,zheng2022weakly2} and (b) our proposed flatten gaussian mask.}
   \label{fig:mask}
\end{figure}
\begin{equation}
\begin{aligned}
\mathbf{m}_{\alpha}^{p}[i] &= \frac{1}{\sqrt{2 \pi}(\mathbf{w}_{\alpha}^{p}/\sigma)}\text{exp}(-\frac{(i/N_{v} - \mathbf{c}_{\alpha}^{p})^{2}}{2(\mathbf{w}_{\alpha}^{p}/\sigma)^{2}}),
\end{aligned}
\end{equation}
where the $\mathbf{m}_{\alpha}^{p} \in \mathbb{R}^{N_{v}}$ is the base mask and  $\mathbf{c}_{\alpha}^{p}$, $\mathbf{w}_{\alpha}^{p}$ denote the center point and width, where a superscript $p \in [1,\cdots,p_{\alpha}]$ denotes $p$-th proposal. The $\sigma$ is a hyperparameter.
The $i \in [1,\cdots,N_{v}]$ denotes the $i$-th index in the length of video. 
We flatten the base mask in the region of proposals by substituting the attention values of it as mean values of them: $\mathbf{m}_{\alpha}^{p}[\textrm{st: ed}] = \textrm{Mean}(\mathbf{m}_{\alpha}^{p}[\textrm{st: ed}])$, where
$\textrm{st} = \mathbf{c}_{\alpha}^{p}-(\frac{\mathbf{w}_{\alpha}^{p}}{2})$, $\textrm{ed} = \mathbf{c}_{\alpha}^{p}+(\frac{\mathbf{w}_{\alpha}^{p}}{2})$, and Mean($\cdot$) is a mean-pooling while keeping the input dimension.
Thus, $\mathbf{m}^{p}_{\alpha}[i]$ is updated as $i$-th value of the Flatten Gaussian function corresponding to $p$-th proposal.
Using $\mathbf{m}^{p}_{\alpha}$, the video features $\mathbf{v} \in \mathbb{R}^{N_{v} \times d}$ are attended to produce $p$-th proposal features:
\begin{equation}
\begin{aligned}
\mathbf{v}^{p}_{\alpha} = \mathbf{v} \circ \mathbf{m}^{p}_{\alpha} \in \mathbb{R}^{N_{v} \times d},
\end{aligned}
\end{equation}
where $\circ$ is column-wise multiplication. 
Therefore $\mathbf{v}^{p}_{\alpha}$ is our final complexity adaptive proposal features.
In the following, we use the $\mathbf{v}^{p}_{\alpha}$ to learn proposal-language alignment by our designed proposal enhancement framework.
% l은 proposal의 개수를 선택할 수 있는 선택지들의 수이다.
%first define minimum and maximum value of the number of proposals % 프로포절의 최소수과 최대수을 준비하고 이 사이에서 프로포절을 결정하도록 하였다.

% 프로포절의 양과 길이
% CPG는 주어진 비디오의 장면 복잡도를 고려하여 proposal을 생성하도록 고안되었다.
% - Number of Proposal
% N ~ M 개의 Proposal에 대해서 선택하도록 Scene Complexity에 따라 선택하도록
% - Length of Proposal
% 
\subsection{Complexity-Adaptive Proposal Enhancement}
Due to the unavailability of supervision, wsVMR systems rely on several training objectives. 
As shown in Figure \ref{fig:model}(d), to enhance the adaptive proposal features $\mathbf{v}^{p}_{\alpha}$, SCANet contains the multiple representation enhancements: (1) Cross-modal Reconstruction and (2) Hierarchical Contrastive Learning, where they are dynamically calibrated according to the scene complexity $\alpha$.

\paragraph{Cross-modal Reconstruction.}
Cross-modal reconstruction aims to learn connections of common semantics among the modalities (\ie video, query), such that we mask a part of the features in one modality and restore that part by referring to the other modality.
Depending on which modality is masked, it is referred to as follows: (1) masked query reconstruction (MQR), and (2) masked video reconstruction (MVR).
For the MQR, we randomly sample verb or noun tokens to mask in query $Q=\{w_{1} \cdots w_{N_{q}}\}$ (\ie $w$ is a word token).
Before masking, we define these tokens as target tokens $w^{tgt}$ to predict, and they are replaced by [mask] tokens in the query, which makes masked query features $\mathbf{q}^{msk} \in \mathbb{R}^{N_{q} \times d}$ throughout text encoder.
Thus a masked query reconstruction loss $\mathcal{L}_{mqr}$ is defined by cross-entropy loss to predict the target words $w^{tgt}$ in the mask from the $\mathbf{q}^{msk}$ and proposal features $\mathbf{v}^{p}_{\alpha}$ as given below:
\begin{equation}
\begin{aligned}
\mathcal{L}_{mqr}(\theta) = -\frac{1}{p_{\alpha}}\sum_{p=1}^{p_{\alpha}}\text{log}f_{\theta}(w^{tgt}|\mathbf{q}^{msk},\mathbf{v}^{p}_{\alpha}),
\end{aligned}
\end{equation}
where $\theta$ is learnable weight and $f_{\theta}$ is a decoder to reconstruct the target words.
For the MVR, we sample video frame features with a probability of 10\%\footnote{Masking is performed in a range of $[\mathbf{c}^{p}_{\alpha}-(\mathbf{w}^{p}_{\alpha}/2),\mathbf{c}^{p}_{\alpha}+(\mathbf{w}^{p}_{\alpha}/2)]$ in video frames assuming effective region of the proposal features $\mathbf{v}^{p}_{\alpha}$.}, where they are defined as $\mathbf{v}^{tgt}$, and then replaced by zeros to make masked proposal features $(\mathbf{v}_{\alpha}^{p})^{msk}$.
As the $\mathbf{v}^{tgt}$ is the $d$-dimensional video features, we introduce a regressor $g_{\theta}$ to regress the features.
Thus, video reconstruction loss $\mathcal{L}_{mvr}$ is defined by L2 loss between target and regressed features as below:
\begin{equation}
\begin{aligned}
\mathcal{L}_{mvr}(\theta) = \frac{1}{p_{\alpha}}\sum_{p=1}^{p_{\alpha}}||\mathbf{v}^{tgt} - g_{\theta}(\mathbf{q},(\mathbf{v}^{p}_{\alpha})^{msk})||_{2}^{2}.
\end{aligned}
\end{equation}
%
\begin{comment}
\begin{figure}[t]
  \centering
  \includegraphics[width=\linewidth]{./figs/adv.pdf}
   \caption{Illustration of adversarial contrastive learning: (a) Training of SCANet and (b) Training of SCANet (Fake).}
   \label{fig:adv}
\end{figure}
\end{comment}
%
\paragraph{Hierarchical Contrastive Learning.}
Contrastive learning aims to enhance the representations of positive features (\ie proposal region) via comparing negative features (\ie non-proposal region).
It is crucial to find a hard negative case (\ie a scene similar to a positive but not a positive).
Therefore, we build hierarchical contrastive learning to explore the hard negatives at the video-level and corpus-level.
For the video-level, we use the input video $V$ and mine the negative in the region of $V$ excluding the adaptive proposal $\mathbf{v}_{\alpha}^{p}$.
% 두가지 
%and there have been effective methods \cite{zheng2022weakly1,zheng2022weakly2} of mining the hard negative in the same video (\ie Intra-video), which has been superior to using different videos (\ie Inter-videos) in a batch.
% 
%SCANet involves both Intra-video and Inter-videos to mine the hard negatives, where we propose an adversarial hard negative mining for Inter-videos, which is comparable to utilizing Intra-video.
%
We first make the negative mask from the positive mask as $1 - \mathbf{m}^{p}_{\alpha}$, and get the negative proposal features $\mathbf{\bar{v}}^{p}_{\alpha} = \mathbf{v} \circ (1 - \mathbf{m}^{p}_{\alpha}) \in \mathbb{R}^{N_{v} \times d}$.
We then, compare the masked query reconstruction losses between positive proposals ($\mathcal{L}_{mqr}$) and negative proposals ($\mathcal{L}_{mqr}^{*}$), which defines video-level contrastive loss $\mathcal{L}_{vid}$ with margin of $\delta_{1}$as:
\begin{equation}
\begin{aligned}
\mathcal{L}_{vid}(\theta) &= \text{max}(\mathcal{L}_{mqr}(\theta)-\mathcal{L}_{mqr}^{*}(\theta) + \delta_{1}, 0),\\
\mathcal{L}_{mqr}^{*}(\theta) &= -\frac{1}{p_{\alpha}}\sum_{p=1}^{p_{\alpha
}}\text{log}f_{\theta}(w^{tgt}|\mathbf{q}^{msk},\mathbf{\bar{v}}^{p}_{\alpha}).
\end{aligned}
\end{equation}
For the corpus-level, we use a video corpus $D_{V}$ composed of all videos in the dataset and mine the hard negative videos for contrastive learning.
To find the hard negatives, we perform video retrieval on the corpus, which takes the query $Q$ and video corpus $D_{V}$ as inputs and predicts the top-k videos as outputs $V_{k} = \textrm{SCANet}^{k}(Q, D_{V})$\footnote{See details about video retrieval of $\textrm{SCANet}^{k}(\cdot,\cdot)$ in Section 5.}, where $V_{k}$ is the top-k videos with the lowest $\mathcal{L}_{mqr}$ for the input query.
As the $V_{k}$ is utilized for negative videos, the ground-truth video is removed from it.
Similar to video-level, we compare the masked query loss between positive proposals ($\mathcal{L}_{mqr}$) and negative videos ($\mathcal{L}_{mqr}^{\dagger}$), which defines the corpus-level contrastive loss $\mathcal{L}_{cps}$ with a margin $\delta_{2}$ below:
%first predict video retrieval on all the videos pertinent to give query.
%
%As shown in Figure \ref{fig:adv}, we first prepare two retrieval models with the same structure: (a) SCANet and (b) SCANet (Fake), where the SCANet is trained from the loss $\mathcal{L}^{r}(\theta) = \mathcal{L}_{mqr}(\theta) + \mathcal{L}_{mvr}(\theta) + \mathcal{L}_{intra}(\theta)$ computed by original video-query pairs, but the SCANet (Fake) is trained from fake loss $\mathcal{L}^{f}(\phi) = \mathcal{L}_{mqr}^{f}(\phi) + \mathcal{L}_{mvr}^{f}(\phi) + \mathcal{L}_{intra}^{f}(\phi)$ computed by top-k negative videos-query pairs.
%
%For the top-k negative videos, they are videos retrieved by trained SCANet\footnote{Refer more details in Model Settings in Section 4.3.} for a given query and multiple videos in dataset as video retrieval task, but the ground-truth video is removed from the top-k retrievals.
%
%Therefore, they are all negative videos but similar with the input query.
%
%By the fake loss $\mathcal{L}^{f}$, the SCANet (Fake) is trained to find the most similar scene in the top-k negative videos for a given query.
%
%Here, we compare the $\mathcal{L}_{mrq}$ in SCANet and SCANet (Fake), which defines the adversarial contrastive loss as:
%
\begin{equation}
\begin{aligned}
\mathcal{L}_{cps}(\theta) &= \text{max}(\mathcal{L}_{mqr}(\theta)-\mathcal{L}_{mqr}^{\dagger}(\theta) + \delta_{2}, 0),\\
\mathcal{L}_{mqr}^{\dagger}(\theta) &= -\frac{1}{k}\sum_{i=1}^{k}\text{log}f_{\theta}(w^{tgt}|\mathbf{q}^{msk},\mathbf{\bar{v}}_{i}),
\end{aligned}
\end{equation}
where $\mathbf{\bar{v}}_{i} \in \mathbb{R}^{N_{v} \times d}$ is $i$-th negative video features from $V_{k}$.
%where $\delta_{2}$ is a margin. The $\phi$ is learnable parameters only trained in the fake model.\footnote{SCANet is optimized by $\theta$ and SCANet (Fake) is optimized by $\phi$ which promotes adversarial training between two models.}
%
%
%
\paragraph{Dynamic calibration.} The videos with high complexity are usually more difficult to train than videos with lower complexity, as the moment predictions in the videos with high complexity are performed under many proposals. Thus we dynamically calibrate the loss to place different weights according to the complexity $\alpha$ as below:
\begin{equation}
\begin{aligned}
\mathcal{L} = \frac{\gamma}{1 + e^{-\alpha}} (\mathcal{L}_{mqr} + \mathcal{L}_{mvr} + \mathcal{L}_{vid} + \mathcal{L}_{cps}),
\end{aligned}
\end{equation}
where $\gamma$ is a hyperparameter.
%
%where,because videos with many scenes are more difficult to learn, so we tried to give weight to these videos.
%
%
%As SCANet is trained to lower the masked query reconstruction loss by the reconstruction and contrastive learning frameworks, in an inference, 
%
For the inference, SCANet predicts the proposal that generates the lowest $\mathcal{L}_{mqr} + \mathcal{L}_{mvr}$ among the proposals.
For the best proposal, the start-end times ([st,ed]) are inferred from the corresponding proposal mask's width $\mathbf{w}^{p}_{\alpha}$ and center $\mathbf{c}^{p}_{\alpha}$ and scaled by the video duration as $[\text{st}, \text{ed}] = [\mathbf{c}^{p}_{\alpha} - \frac{\mathbf{w}^{p}_{\alpha}}{2}, \mathbf{c}^{p}_{\alpha} + \frac{\mathbf{w}^{p}_{\alpha}}{2}] * \text{duration}$
%
%\begin{equation}
%\begin{aligned}
%[\text{st}, \text{ed}] = [\mathbf{c}^{p}_{\alpha} - \frac{\mathbf{w}^{p}_{\alpha}}{2}, \mathbf{c}^{p}_{\alpha} + \frac{\mathbf{w}^{p}_{\alpha}}{2}] * \text{duration}
%\end{aligned}
%\end{equation}
%
%
\section{Experiments}
\subsection{Dataset}
Our proposed SCANet is validated on three moment retrieval benchmark datasets, where the wsVMR system uses temporal annotations only for evaluation.
\paragraph{Charades-STA.} Charades-STA includes about 30 seconds of videos for human behaviors and their language queries. Average length is about 29.8 seconds, and dataset contains 12,408 video-query pairs and 3,720 for testing. 
%For the weakly-supervised VMR, the temporal annotations are only used for evaluations.
%
\paragraph{ActivityNet Captions.} ActivityNet Captions is a large-scale dataset including about 117 seconds videos of human actions and their language query. The dataset contains 19,290 videos with 37,417/17,505/17,031 smaples for train/val\_1/val\_2 splits.
%
%The temporal annotations are used for evaluation like Charades-STA.
%
SCANet is validated on the val\_2.
\paragraph{\bf{TV show Retrieval.}} TV show Retrieval (TVR) \cite{lei2020tvr} comprises 6 TV shows about diverse genres, including 109K queries from 21.8K multi-character videos with subtitles. Each video is about 60-90 seconds. The TVR is split into $80\%$ train, $10\%$ val, $10\%$ test-public. The test-public is prepared for the challenge. As test-public is currently unavailable, SCANet is validated on the val.
\subsection{Evaluation Metric} To evaluate the moment retrieval, we compute the average recall (R@n) over all queries, where temporal Intersection over Union (IoU=m) measures the overlap between prediction and ground-truth. The n denotes the recall rate of top-n predictions, and m is the predefined IoU threshold, thus quantifying the percentage of predicted moments with the IoU value larger than m among top-n predictions.
\begin{comment}
\subsection{Implementation Details}
\paragraph{Data Settings.} 
%
For the video encoder, I3D \cite{carreira2017quo} model is used to get the Charades-STA video features, and C3D \cite{tran2015learning} model is used for the ActivityNet-Caption video features.
%
Both video features are extracted by every 8 frames.
%
For the word token embedding, we use word2vec from GloVe \cite{pennington2014glove}.
%
The size of the vocabulary is fixed as 8000 with maximum 20 word-length of sentence.
%
\paragraph{Model Settings.} Hyperparameters in SCANet are as follows: $K=12$ for the maximum number of scene complexity, Minumum number of proposal p_{min} = 6, maximum number of proposals p_{max} = 9
$\sigma$ for Gaussian function is $8$, the margins $\delta_{1}$ for contrastive loss $\mathcal{L}_{vid}$ and $\mathcal{L}_{cps}$ are $\delta_{1}=0.1, \delta_{2}=0.5$, where the higher margin of $\delta_{2}$ is designed for distinguishing similar scene from other videos.
% $\gamma$ is 1.2 in equation 12
%
The minimum and maximum number of proposals are $n=5$ and $m=10$.
%
To prepare the ranked top-k (k=15) videos used in SCANet (Fake), they are retrieved by SCANet trained under $\mathcal{L}^{r}$ as performing video retrieval task, where the prediction is also performed by the lowest $\mathcal{L}_{mqr}$ among videos of dataset.
\end{comment}
\begin{table}
  \centering
  \footnotesize
  \setlength{\tabcolsep}{5pt}
  \caption{Performances of weakly-supervised video moment retrieval on the Charades-STA dataset.}
  \begin{tabular}{@{}l|ccc|ccc@{}}
    \toprule
    \multirow{2}{*}{Method} & \multicolumn{3}{c|}{R@1,IoU=m} & \multicolumn{3}{c}{R@5,IoU=m} \\     
      &m=0.3 &m=0.5 &m=0.7 &m=0.3 &m=0.5 &m=0.7 \\
    \midrule
    TGA \cite{mithun2019weakly}         &32.14 &19.94 &8.84   &86.58 &65.52 &33.51\\
    CTF \cite{chen2020look}             &39.80 &27.30 &12.90  &-     &-     &-\\
    SCN \cite{lin2020weakly}            &42.96 &23.58 &9.97   &95.56 &71.80 &38.87\\
    WSTAN \cite{wang2021weakly}         &43.39 &29.35 &12.28  &93.04 &76.13 &41.53\\
    BAR \cite{wu2020reinforcement}      &44.97 &27.04 &12.23  &-     &-     &-\\
    LoGAN \cite{tan2021logan}           &48.04 &31.74 &13.71  &89.01 &72.17 &37.58\\
    MARN \cite{song2020weakly}          &48.55 &31.94 &14.81  &90.70 &70.00 &37.40\\
    WSRA \cite{fang2020weak}            &50.13 &31.20 &11.01  &86.75 &70.50 &39.02\\
    CCL \cite{zhang2020counterfactual}  &-     &33.21 &15.68  &-     &73.50 &41.87\\
    CRM \cite{huang2021cross}           &53.66 &34.76 &16.37  &-     &-     &-\\
    VCA \cite{wang2021visual}           &58.58 &38.13 &19.57  &98.08 &78.75 &37.75\\
    LCNet \cite{yang2021local}          &59.60 &39.19 &18.87  &94.78 &80.56 &45.24\\
    RTBPN \cite{zhang2020regularized}   &60.04 &32.36 &13.24  &97.48 &71.85 &41.18\\
    CNM \cite{zheng2022weakly1}         &60.39 &35.43 &15.45  &-     &-     &-\\
    CPL \cite{zheng2022weakly2}         &65.99 &49.05 &22.61  &96.99 &84.71 &52.37\\
    \midrule[0.1pt]
    SCANet (ours)                              &\textbf{68.04} &\textbf{50.85} &\textbf{24.07} &\textbf{98.24} &\textbf{86.32} &\textbf{53.28}\\
    \bottomrule
  \end{tabular}
  \label{tab:charades_sta}
\end{table}
\begin{table}
  \centering
  \footnotesize
  \setlength{\tabcolsep}{5pt}
  \caption{Performances of weakly-supervised video moment retrieval on the ActivityNet Captions dataset.}
  \begin{tabular}{@{}l|ccc|ccc@{}}
    \toprule
    \multirow{2}{*}{Method} & \multicolumn{3}{c|}{R@1,IoU=m} & \multicolumn{3}{c}{R@5,IoU=m} \\     
      &m=0.1 &m=0.3 &m=0.5 &m=0.1 &m=0.3 &m=0.5 \\
    \midrule
    WS-DEC \cite{duan2018weakly}        &62.71 &41.98 &23.34 &-     &-     &-\\
    EC-SL \cite{chen2021towards}        &68.48 &44.29 &24.16 &-     &-     &-\\
    MARN \cite{song2020weakly}          &-     &47.01 &29.95 &-     &72.02 &57.49\\
    SCN \cite{lin2020weakly}            &71.48 &47.23 &29.22 &90.88 &71.56 &55.69\\
    BAR \cite{wu2020reinforcement}      &-     &49.03 &30.73 &-     &-     &-\\
    RTBPN \cite{zhang2020regularized}   &73.73 &49.77 &29.63 &93.89 &79.89 &60.56\\
    CTF \cite{chen2020look}             &74.20 &44.30 &23.60 &-     &-     &-\\
    WSLLN \cite{gao2020weakly}          &75.40 &42.80 &22.70 &-     &-     &-\\
    LCNet \cite{yang2021local}          &78.58 &48.49 &26.33 &93.95 &\textbf{82.51} &62.66\\
    CCL \cite{zhang2020counterfactual}  &-     &50.12 &31.07 &-     &77.36 &61.29\\
    WSTAN \cite{wang2021weakly}         &79.78 &52.45 &30.01 &93.15 &79.38 &63.42\\
    CRM \cite{huang2021cross}           &81.61 &55.26 &32.19 &-     &-     &-\\
    CNM \cite{zheng2022weakly1}         &78.13 &55.68 &\textbf{33.33} &-     &-     &-\\
    CPL \cite{zheng2022weakly2}         &82.55 &55.73 &31.37 &87.24 &63.05 &43.13\\
    \midrule[0.1pt]
    SCANet (ours) &\textbf{83.62} &\textbf{56.07} &31.52 &\textbf{94.36} &82.34 &\textbf{64.09}\\
    \bottomrule
  \end{tabular}
  \label{tab:activitynet}
\end{table}
\begin{figure*}[t]
  \centering
  \includegraphics[width=\linewidth]{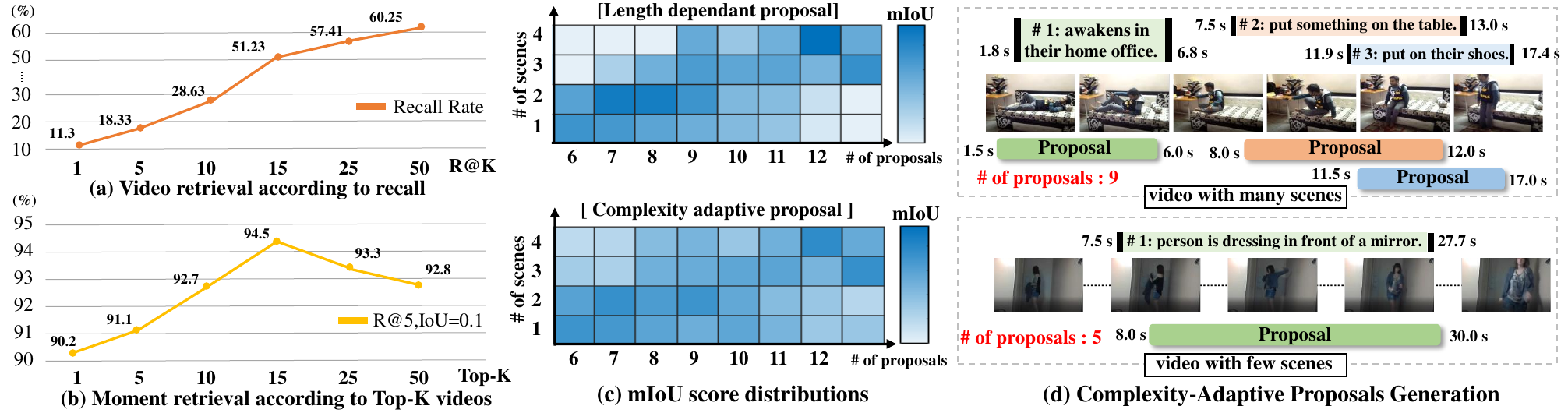}
   \caption{Qualitative results of SCANet: (a) shows the video retrieval performance of SCANet according to the R@k, (b) shows the moment retrieval performances according to involving top-K retrieved videos for hierarchical contrastive learning, (c) shows the IoU scores distributions according to the number of scenes and proposals (upper: length dependant proposals, below: complexity adaptive proposals), and (d) illustrates the proposals in SCANet according to videos with diverse scenes.}
   \label{fig:vr}
\end{figure*}
\subsection{Experimental Results}
Table \ref{tab:charades_sta} and Table \ref{tab:activitynet} summarize the results on Charades-STA (C-STA) and ActivityNet Captions (ANC) datasets.
SCANet is compared to previous works (Please, refer to Related Works for their detailed descriptions). 
SCANet shows the best performances of all metrics on C-STA, which is especially effective in metrics (R@1) and also the effectiveness in four metrics on ANC.
Table \ref{tab:tvr} firstly summarizes the results of wsVMR systems on TVR.
We reproduce the baseline and most recent model from their public codes.
%For the ActivityNet-Caption in Table \ref{tab:activitynet}, SCANet also shows the effectiveness.
%
The videos in TVR are quite challenging because they include relatively more similar actions and backgrounds, which are difficult for the models to distinguish.
Overall, SCANet shows improvements in retrieval quality, but it is also notable that those improvements are mainly from rectifying the video samples suffering scene-proposal mismatch problems, which can be confirmed in Figure \ref{fig:vr}(c).
%Considering the fact that the performance gains in previous works are getting smaller, 
%four metrics in the ANC.
%
%
%
%From the fact that their performance gains are getting smaller, especially in Recall at 1 (R@1) measures, we can also see the difficulties of ANC dataset, where it tells us that to solve a long tailed problem (\ie Handling about specific samples that are present in a small proportion in the dataset and show low accuracy.) in VMR is another future work for video searching technologies.
%
\begin{table}
  \centering
  \footnotesize
  \setlength{\tabcolsep}{5pt}
  \caption{Performances of weakly-supervised video moment retrieval on the TVR dataset (validation) ($\star$ reproduced).}
  \begin{tabular}{@{}l|ccc|ccc@{}}
    \toprule
    \multirow{2}{*}{Method} & \multicolumn{3}{c|}{R@1,IoU=m} & \multicolumn{3}{c}{R@5,IoU=m} \\     
      &m=0.1 &m=0.3 &m=0.5 &m=0.1 &m=0.3 &m=0.5 \\
    \midrule
    $\textrm{TGA}^{\star}$ \cite{mithun2019weakly}        &17.61 &2.38 &0.97 &48.63     &11.54     &5.32\\
    $\textrm{CPL}^{\star}$ \cite{zheng2022weakly2}        &33.16 &7.28 &2.11 &64.41  &17.93     &8.56\\
    \midrule[0.1pt]
    SCANet (ours) &\textbf{37.51} &\textbf{10.76} &\textbf{4.24} &\textbf{67.47} &\textbf{20.32} &\textbf{10.21}\\
    \bottomrule
  \end{tabular}
  \label{tab:tvr}
\end{table}
\begin{table}
  \centering
  \footnotesize
  \caption{Ablation study of redundancy removal for scene complexity estimation ($f_{sc}$) along the types to find the redundancy among queries in the annotated query set $Q_{V}$.}
  \begin{tabular}{@{}c|ccc|c@{}}
    \toprule
    \multirow{2}{*}{Types} & \multicolumn{3}{c|}{R@1,IoU=m} & \multicolumn{1}{c}{R@5,IoU=m} \\     
       &m=0.1 &m=0.3 &m=0.5 &m=0.1 \\
    \midrule
       %\multicolumn{2}{c|}{Baseline (no-use)}    &75.31 &27.16 &50.98      &88.43\\\midrule
     none                               &77.43  &49.78 &27.32     &88.32\\
     noun                               &83.54  &55.97 &31.82     &94.50\\
     verb                               &81.26  &52.32 &30.91     &93.11\\
     noun \& verb                       &80.52  &51.31 &30.42     &92.52\\
    \bottomrule
  \end{tabular}
  \label{tab:ablation_sce}
\end{table}
\subsection{Ablation Study}
Ablation studies are performed on the ActivityNet Captions (validation)\footnote{The validation set of Charades-STA is not available.}.
To show performance variances in various metrics, we validate SCANet on the most challenging metric (R@1,IoU=0.5) and the easiest one (R@5,IoU=0.1).
Table \ref{tab:ablation_sce} summarizes the studies about redundancy removal in the annotated query set $Q_{V}$ for scene complexity estimation ($f_{sc}$).
%
%We first estimate our base without scene complexity, where the number of proposals is fixed as 8. 
%
%the scene complexity $\alpha$ from the video length by giving high value ($\alpha$=6) for the long-length video and a small value ($\alpha$=1) for the short-length video, which implies spurious correlations between video-length and proposal generations.
%
%
The first section is the results of complexity estimation without redundancy removal, where the complexity equals the number of queries in $Q_{V}$.
The below sections are the results with redundancy removal, where the noun or verb is used to find the redundant queries.
Redundancy removal with the noun or verb shows effectiveness. 
However, using both together decreases the performance.
We consider that finding redundant queries using noun and verb together can make sure to find almost the same descriptions (\eg `person \textit{drinks} a cup of \textit{coffee}' and `person \textit{drinks} \textit{coffee} in table'), but the redundancy shows variance for the same scene by changing noun or verb (\eg `one \textit{holds} a cup of coffee').
\begin{table}
  \centering
  \footnotesize
  \caption{Ablation study of generating proposals. (n: number of proposals), (w, s: frame width and stride of window), (G: Gaussian mask, FG: Flatten Gaussian mask)}
  \begin{tabular}{@{}l|cc|c@{}}
    \toprule
    \multirow{2}{*}{Method} & \multicolumn{2}{c|}{R@1,IoU=m} & \multicolumn{1}{c}{R@5,IoU=m} \\     
       &m=0.1 &m=0.5 &m=0.1 \\
    \midrule
       %\multicolumn{2}{c|}{Baseline (no-use)}    &75.31 &27.16 &50.98      &88.43\\\midrule
     fixed proposal (n:6)                         &77.43  &27.32      &88.32\\
     fixed proposal (n:8)                         &78.23  &27.98      &90.12\\\midrule
     sliding window (w:\{20,40\},s:5)                 &68.54  &26.51      &84.32\\
     sliding window (w:\{20,40,60\},s:10)              &69.14  &24.33      &86.72\\\midrule
     complexity adaptive proposal (G)                           &81.32  &29.43      &92.41\\
     complexity adaptive proposal (FG)                           &83.54  &31.82      &94.50\\
    \bottomrule
  \end{tabular}
  \label{tab:proposal}
\end{table}
Table \ref{tab:proposal} shows the ablation studies of proposal generations in SCANet.
Our baseline was to generate proposals with fixed numbers, and here, to determine the locations and lengths of the proposals, we use a learnable $d$-dimensional single vector instead of complexity features $\mathbf{Z}_{\alpha} \in \mathbb{R}^{d}$.
We modify the baseline with the sliding window and complexity adaptive proposals, where the adaptive method was the most effective with our designed Flatten Gaussian mask.
Table \ref{tab:ablation_loss} shows the ablation studies with two proposal enhancements and their calibrating method.
Incremental improvements are shown by adding two enhancements and calibrating the enhancements with complexity.
%where $\mathcal{L}_{adv}$ shows huge gain, which tells that the effective hard negatives are also in inter-videos, but it also has a disadvantage by using more memory resources from the fake model.
%
\begin{table}
  \centering
  \footnotesize
  \caption{Ablation study in Complexity-Adaptive Proposal Enhancement. (CMR: cross-modal reconstruction, HCL: hierarchical contrastive learning, Calibration: calibrating training loss according to scene complexity $\alpha$).}
  \begin{tabular}{@{}ccc|c|c@{}}
    \toprule
    \multicolumn{3}{c|}{Proposal Enhancement} & \multicolumn{1}{c|}{R@1,IoU=m} & \multicolumn{1}{c}{R@5,IoU=m} \\
      CMR & HCL & Calibration &m=0.5 &m=0.1 \\
    \midrule
    \checkmark  &            &               &28.45  &91.52\\
    \checkmark  & \checkmark &               &30.32  &92.71\\
    \checkmark  & \checkmark & \checkmark    &31.82  &94.50\\
    \bottomrule
  \end{tabular}
  \label{tab:ablation_loss}
\end{table}

Figure \ref{fig:vr}(a) shows video retrieval performances of SCANet that predicts recall rate with top-k videos (R@K), and (b) shows the moment retrieval performances according to using top-k videos for $\mathcal{L}_{cps}$.
As K increases, the recall increases in (a), and the moment retrieval performance in (b) is also improved using the predicted videos as negative (\ie ground-truth videos are removed).
Meanwhile, over K=15, the performance degrades, where we presume that the top-k videos start to contain not hard negative videos.
%
%Figure \ref{fig:vr} (c) shows the performances gain when using negative video for inter-videos as top-k videos than videos in a batch, which explains the effectiveness of introduction of top-k videos from video retrieval.
%
%
\subsection{Qualitative Results}
Figure \ref{fig:vr}(c) shows the retrieval performances according to the number of proposals and the number of scenes for each video.
Compared to the length-dependant proposals (\eg sliding window), complexity adaptive proposals mitigates the scene-proposal mismatch by regularizing the retrieval quality along the number of proposals and scenes.
Figure \ref{fig:vr}(d) illustrates the proposals of SCANet on (a) short-length video with many scenes and (b) long-length video with few scene.
The adaptive proposals accurately capture the scenes in both cases.
In the bottom left, we also add the number of generated proposals for each video, which shows that proposals do not depend on the length of the video.
%

%In Figure \ref{fig:qual} (b), we also shows the other ranked proposals, and they also have high-overlap with GT. 
%
%Please, refer more results represented in the appendix.
%
%
\begin{comment}
\begin{figure}[t]
  \centering
  \includegraphics[width=0.95\linewidth]{./figs/qual.pdf}
   \caption{Illustration of moment predictions on (a) short-length video with many scenes and (b) long-length video with few scenes.}
   \label{fig:qual}
\end{figure}
\end{comment}
%
\section{Implementation Details}
\paragraph{Data Settings.} 
For the video encoder, I3D \cite{carreira2017quo} model is used to get the Charades-STA video features, and C3D \cite{tran2015learning} model is used for the ActivityNet-Caption video features.
Both video features are extracted by every 8 frames.
For the word token embedding, we use word2vec from GloVe \cite{pennington2014glove}.
The size of the vocabulary is fixed as 8000 with maximum 20 word-length of sentence.
\paragraph{Model Settings.} Hyperparameters in SCANet are as follows: $K=12$ for the maximum number of scene complexity, the minimum number of proposal $p_{\textrm{min}}$ = 5, the maximum number of proposals $p_{\textrm{max}}$ = 14, the hyperparameter of calibration is $\gamma=0.5$.
The $\sigma$ for Gaussian function is $8$, the margins $\delta_{1}$ for contrastive loss $\mathcal{L}_{vid}$ and $\mathcal{L}_{cps}$ are $\delta_{1}=0.1, \delta_{2}=0.5$, where the higher margin of $\delta_{2}$ is designed for promoting to distinguish the positive video from negative videos with similar scenes.
% $\gamma$ is 1.2 in equation 12
%
\paragraph{Video Retrieval with SCANet.}
To prepare the ranked top-k videos used in $V_{k} = \textrm{SCANet}^{k}(Q, D_{V})$, they are retrieved by SCANet trained from reconstruction losses (\ie $\mathcal{L}_{mqr},\mathcal{L}_{mvr}$) and video-level contrastive loss (\ie $\mathcal{L}_{vid}$). SCANet retrieves top-k (\eg k=15) videos from the video dataset for each query that have the lowest reconstruction losses.
To train wsVMR, the top-k video IDs are utilized to provide hard negative videos, which makes $\mathcal{L}_{cps}$.
The retrieval performances (\ie R@K, prediction recall according to top-K videos) are presented in Figure \ref{fig:vr}.
\begin{figure}[t]
  \centering
  \includegraphics[width=\linewidth]{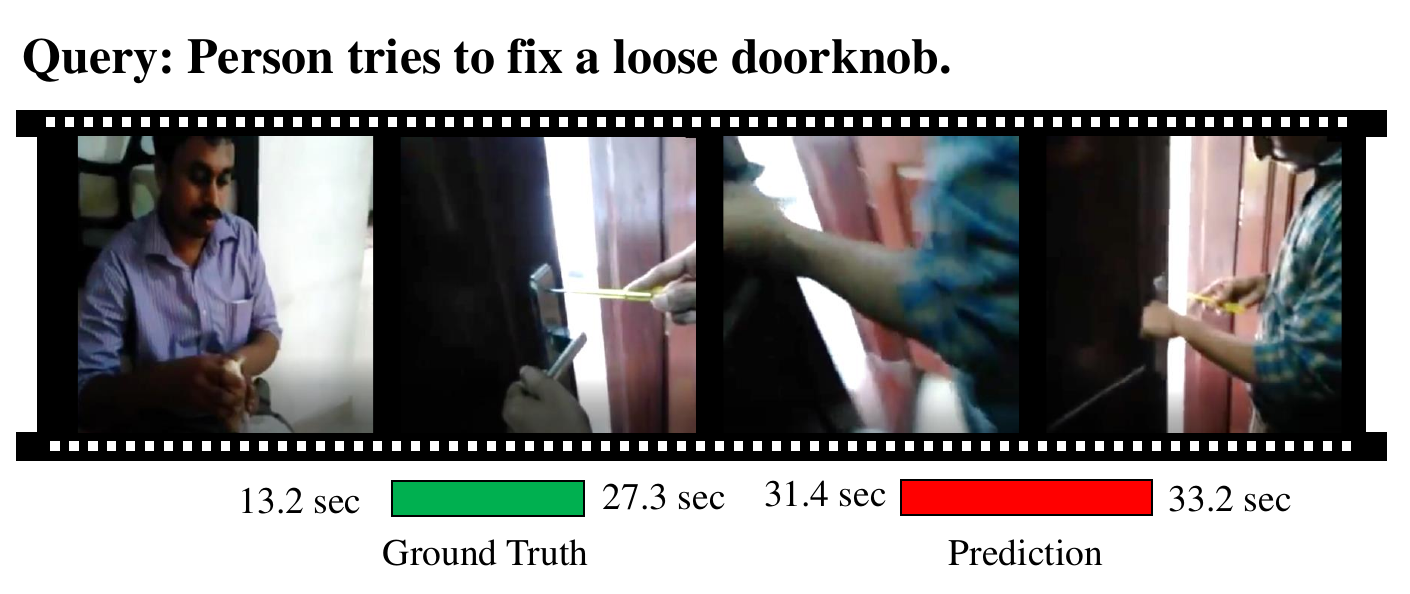}
   \caption{Illustration of failure case of moment prediction.}
   \label{fig:fail}
\end{figure}
\section{Failure cases}
Figure \ref{fig:fail} presents the failure case of our proposed SCANet.
For the query `Person tries to fix a doorknob', SCANet predicts the moment of person openning the door.
%
%We identify that other wVMR systems also share the common failure cases. 
%
We consider the action of fixing something is not frequently paired with queries, and it may be unavailable for wsVMR systems to directly learn about the uncommon video-query pairs as a long-tailed action recognition problem.
Therefore, wsVMR systems are more vulnerable to these actions with categories in the long tail and we believe that overcoming the long-tail problem should be a contribution to many tasks including moment retrieval. 
%as they can not aided from supervision of boundary information.
% 이것은 long-tailed action recognition에 대한 문제와 관련있으며, wVMR system들 원리적으로 이 문제에 취약할 수 밖에 없다.
%
%
Our future works also include mitigating this long-tail problem.

\section{Limitations}
Our proposed method is based on the scene complexity of video by referring to the number of annotated queries to the video, which can also have more flexibility by applying other systems \cite{paul2018w,huang2022weakly} under weak supervision.
However, in the real environment, it may not be available to get scene complexity of video from referring to other annotated queries (\ie In real environment, we may not access to other annotated query sets for one video).
We feel this is our current SCANet's limitation, and to overcome this, we further made another effort to learn scene complexity via the neural network from the input of video, where we refer to this method as `Scene Complexity Neural Estimator'. In our supplementary materials, we elaborate on this with our current studies as another our experimental contributions.

\section{Conclusion}
SCANet is presented to consider a scene-proposal mismatch problem in the wsVMR.
%, where current proposals do not properly capture scenes in a video.
%
SCANet measures a scene complexity of multiple scenes in each video.
Founded on the complexity, SCANet builds complexity-adaptive proposal generation to mitigate the scene-proposal mismatch and complexity-adaptive proposal enhancement to enhance the representation by calibrating with the complexity.
%
%
%We discuss limitations and potential negative effects of our work in appendix.
\section*{Acknowledgment}
This work was supported by Institute for Information \& communications Technology Planning \& Evaluation (IITP) grant funded by the Korea government(MSIT) (No. 2021-0-01381, Development of Causal AI through Video Understanding and Reinforcement Learning, and Its Applications to Real Environments) and partly supported by the National Research Foundation of Korea (NRF) grant funded by the Korea government(MSIT) (No. 2022R1A2C2012706).

{\small
\bibliographystyle{ieee_fullname}
\bibliography{egbib}
}

\end{document}